\documentclass[letterpaper]{article} 
\usepackage{aaai2026}  
\usepackage{times}  
\usepackage{helvet}  
\usepackage{courier}  
\usepackage[hyphens]{url}  
\usepackage{graphicx} 
\usepackage{natbib}  
\usepackage{caption} 
\usepackage{algorithm}
\usepackage{algorithmic}

\usepackage{newfloat}
\usepackage{listings}
\DeclareCaptionStyle{ruled}{labelfont=normalfont,labelsep=colon,strut=off} 
\floatstyle{ruled}
\newfloat{listing}{tb}{lst}{}
\floatname{listing}{Listing}
\title{Position on LLM-Assisted Peer Review: Addressing Reviewer Gap \\ through Mentoring and Feedback}
\author{
    JungMin Yun\equalcontrib \textsuperscript{\rm 1},
    JuneHyoung Kwon\equalcontrib \textsuperscript{\rm 1},
    MiHyeon Kim\equalcontrib \textsuperscript{\rm 3}, 
    YoungBin Kim\textsuperscript{\rm 1, 2}
}
\affiliations{
    \textsuperscript{\rm 1} Department of Artificial Intelligence, Chung-Ang University\\
    \textsuperscript{\rm 2} Graduate School of Advanced Imaging Sciences, Multimedia and Film, Chung-Ang University\\
    \textsuperscript{\rm 3} KT Corporation\\
    cocoro357@cau.ac.kr, dirchdmltnv@cau.ac.kr, mihyeon.gim@kt.com, ybkim85@cau.ac.kr
}

\usepackage{bibentry}

\begin{document}

\maketitle

\begin{abstract}
The rapid expansion of AI research has intensified the Reviewer Gap, threatening the peer-review sustainability and perpetuating a cycle of low-quality evaluations. This position paper critiques existing LLM approaches that automatically generate reviews and argues for a paradigm shift that positions LLMs as tools for assisting and educating human reviewers. We define the core principles of high-quality peer review and propose two complementary systems grounded in these foundations: (i) an LLM-assisted mentoring system that cultivates reviewers’ long-term competencies, and (ii) an LLM-assisted feedback system that helps reviewers refine the quality of their reviews. This human-centered approach aims to strengthen reviewer expertise and contribute to building a more sustainable scholarly ecosystem.

\end{abstract}


\section{Introduction}

High-quality peer review is the foundation that sustains the credibility and scholarly authority of academic conferences. Recent growth in AI research has expanded at an unprecedented pace, rapidly exceeding the processing capacity of existing peer review systems. The resulting strain extends beyond individual reviewer workload and now imposes structural pressure on the entire conference ecosystem~\cite{schaeffer2025position}. This pressure arises from a widening imbalance between the rapid expansion of the AI field and the relatively stagnant growth of the review infrastructure. We refer to this fundamental mismatch as the Reviewer Gap, which comprises two mutually reinforcing components.

\textbf{Volume Gap (Capacity Constraints).} Submission volumes in AI are rising at unprecedented levels. Conferences such as ICLR and ACL report annual increases of 20–30\%~\cite{cspaper2025iclrstorm}, and NeurIPS received more than 27,000 submissions in 2025, representing approximately a tenfold increase over the past decade~\cite{wang2025neuripssubmissions}. This trend structurally exceeds conference review capacity~\cite{tran2020open}, resulting in excessive assignments per reviewer. The resulting review fatigue leads to superficial commentary, insufficiently justified critiques, and perfunctory assessments, collectively degrading review quality~\cite{wei2025ai, adam2025peer, schaeffer2025position}.

\textbf{Quality Gap (Expertise Constraints).} Submission growth outpaces the increase in domain experts, including faculty members and industry researchers~\cite{schaeffer2025position}. To address this imbalance, conferences have expanded reviewer pools by recruiting junior reviewers with limited publication and reviewing experience or by mandating author participation, as seen in venues such as ARR and NeurIPS. Although such strategies enable quantitative expansion, they exacerbate disparities in expertise. Without systematic training, these conditions risk entrenching inexperienced practices and structurally reproducing low-quality review norms~\cite{galipeau2013systematic, kwglobal2025reviewertraining, dewitt2001bad,  aczel2025present}.

These two gaps reinforce one another,  creating a vicious cycle of low-quality reviewing. Poor-quality reviews fail to provide constructive feedback, hindering research progress and undermining both conference credibility and researcher motivation. Existing Large Language Model (LLM) approaches have primarily focused on automated review generation and summarization, but these methods neither reliably ensure quality nor support long-term reviewer development~\cite{kocak2025ensuring, lee2025role}. Addressing the Reviewer Gap, therefore, is not simply a matter of generating more reviews faster. The central challenge lies in systematically cultivating reviewer competence and structurally improving review quality.

Building on this perspective, this position paper proposes an LLM-assisted reviewer mentoring and feedback system that shifts the focus from automated review generation to a more human-centered approach. We introduce a unified rubric defining core principles of high-quality reviews and present a dual-structured framework comprising a mentoring-oriented educational system and a feedback-driven verification system. This framework repositions LLMs as educational assistants that support reviewer capability enhancement. Furthermore, this approach has the potential to catalyze a long-term virtuous cycle in which better reviews lead to refined research outputs, which advance AI technologies that further strengthen the review ecosystem, ultimately enhancing the credibility and expertise of the scholarly community.

\begin{figure*}[t!]
    \centering
    \includegraphics[width=0.99\textwidth]{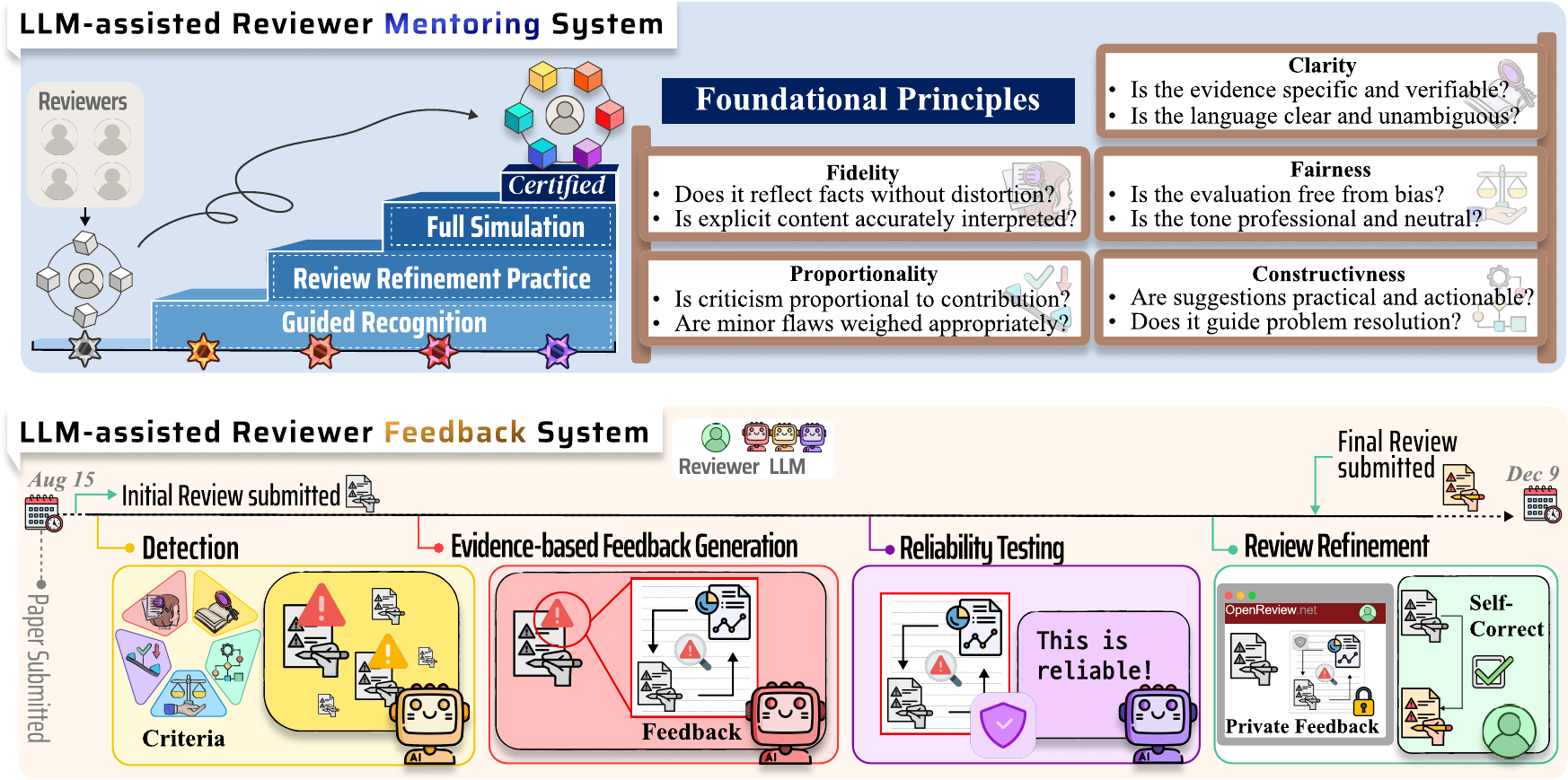}
    \caption{Overview of our proposed LLM-assisted reviewer mentoring and feedback system. The framework integrates foundational principles of high-quality reviews with a dual-system architecture comprising the mentoring system and the feedback system.}
    \label{fig:overview}
\end{figure*}

\section{Diagnosis of the Limitations in LLM-based Peer Review Systems}
Recent efforts have integrated LLMs into the review process at several AI conferences. For instance, AAAI experimentally introduced a reasoning model-based AI review system~\cite{aaai2025aipilotfaq}, while ICLR encouraged reviewers to refine their drafted comments using LLM-generated suggestions aimed at improving clarity and actionability~\cite{thakkar2025can}. Although these initiatives have contributed to alleviating review burden to some extent, they exhibit limitations from both technical and structural perspectives.

The most significant concern arises when LLMs directly generate initial reviews. This approach entails several risks~\cite{chitnis2024autoref}. These include the generation of erroneous information due to hallucination, inaccurate understanding of research contributions and scholarly context, and superficial summarization from excessive generalization~\cite{xu2025can, kocak2025ensuring, pataranutaporn2025can}. Reviewing demands expert judgment, including the assessment of contributions, contextualization within the literature, and evaluation of methodological validity. Current LLMs have limitations in fully substituting for such expertise~\cite{szymanski2025limitations}. The problem is compounded when inexperienced reviewers rely on LLM-generated drafts without critical examination, potentially reproducing misunderstandings and biases.

These concerns suggest that the value of LLMs in peer review lies not in replacing human judgment but in supporting it. A more effective alternative is to use LLMs not as autonomous review generators but as tools providing feedback on human-written reviews~\cite{chen2025mlr, kamoi2024evaluating, xu2025can, zhao2025abgen}. Human reviewers maintain responsibility for substantive, domain-informed evaluation, while LLMs analyze draft reviews to identify unclear reasoning, insufficient justification, or structural issues. This represents a balanced review workflow that preserves expert judgment while leveraging LLM strengths in linguistic refinement and consistency checking~\cite{aaai2025aipreviewsystem, thakkar2025can, naddaf2025aitransformpeerreview}.

However, this feedback-oriented approach has limitations if implemented as a one-time intervention. Without a sustainable educational framework, reviewers may accept incomplete or biased suggestions without critical examination, reinforcing low-quality review practices~\cite{kocak2025ensuring, zhu2025your}. To mitigate this risk, LLM-based assistance should evolve into a long-term training mechanism that enhances reviewer capabilities through repeated engagement and progressive skill development.

Building on these insights, we propose a paradigm that positions LLMs as feedback and educational systems for human reviewers. This paradigm augments reviewer competence through two complementary functions that address both the quantitative and qualitative dimensions of the Reviewer Gap (Figure~\ref{fig:overview}): (i) systematic instruction to help reviewers internalize core components of high-quality peer review, and (ii) support for identifying and correcting potential errors or biases before submission. Together, these mechanisms address the immediate need for review quality improvement while advancing the long-term goal of cultivating a more capable reviewer community.

\section{Foundational Principles for Review Quality}
For the proposed LLM-assisted reviewer mentoring and feedback system to function effectively and consistently, it is essential to define quality criteria that reviews should satisfy. The absence of standardized criteria for high-quality reviews obscures the reference points that LLMs need to consistently evaluate and support the improvement of reviews. To address this issue, we propose five core principles that form the foundation of high-quality reviews by synthesizing the key elements commonly required across major conferences~\cite{aaai2025reviewerinstructions, arr2025reviewerguidelines, neurips2025reviewerguidelines}. These principles serve as an evaluation rubric for LLM-based mentoring and review improvement feedback.

\begin{itemize}
    \item \textbf{Fidelity:} Reviews must accurately reflect the factual information, core claims, and interpretations of experimental results presented in the paper without distortion. This includes verifying that reviewers have not overlooked or misunderstood content explicitly stated in the manuscript.
    \item \textbf{Clarity:} Reviews should be written in clear and unambiguous language, with all claims supported by specific and verifiable evidence. Vague statements or unfounded criticisms fail to help authors identify problems and derive directions for improvement.
    \item \textbf{Fairness:} Reviews must evaluate papers based on the scholarly contribution and validity, free from bias against researchers, institutions, or research areas. It is essential to exclude emotional or dogmatic expressions and maintain a professional and neutral tone for legitimate and consistent evaluation.
    \item \textbf{Proportionality:} The intensity and focus of criticism should be proportional to the paper's core contribution and soundness. It is inappropriate to focus excessively on minor flaws or to overinterpret localized weaknesses as fatal flaws that negate the entire value of the research. Reviewers should calibrate critique weight commensurate with significance.
    \item \textbf{Constructiveness:} High-quality reviews should include practical, actionable suggestions for improvement, not merely identify flaws. This aligns with the view that peer review should focus on improvement, contributing to knowledge advancement within the scholarly community.
\end{itemize}


These five core principles provide a multidimensional evaluation framework for improving review quality and consistently align the design and functionality of LLM-based reviewer support systems. The educational and feedback mechanisms proposed in this work aim to systematically enhance review quality based on these principles and strengthen the reliability and sustainability of the peer review ecosystem. Ultimately, these guidelines embody the core philosophy of our system, which seeks to cultivate better reviewers, not merely to produce better reviews.

\section{LLM-Assisted Reviewer Mentoring System}
The structural causes of the Reviewer Gap extend beyond a mere shortage of reviewers to encompass the absence of educational mechanisms through which reviewers can systematically acquire the competencies required for high-quality evaluation. While a training-based foundation is necessary to strengthen reviewer expertise over the long term, mandatory training programs risk undermining reviewer autonomy and motivation. Moreover, conferences face substantial practical constraints in operating such programs due to limitations in cost, staffing, and scalability. Given these challenges, we argue that a support-oriented, voluntary `safe-to-fail' learning ecosystem represents a more sustainable and desirable approach.

Recent advances in LLM research provide a technological foundation capable of addressing this educational need. Studies have empirically demonstrated that LLMs can analyze learner response patterns to automatically diagnose weaknesses and generate adaptive curricula, personalized practice problems, and explanation-based feedback tailored to individual proficiency levels~\cite{wen2024ai, li2025adaptive, luo2025assessing}. This offers a promising pathway to simultaneously achieve the personalization, scalability, and continuous improvement required for reviewer development. In other words, LLMs are not intended to replace reviewers; rather, they can function as educational assistants that help reviewers progressively internalize core principles throughout the review process, thereby contributing to the mitigation of the Reviewer Gap.

Based on this perspective, we propose an LLM-assisted reviewer mentoring system grounded in a pedagogically structured curriculum. Rather than immediately demanding complex review-writing abilities, this system features a step-by-step learning structure designed to help learners progressively internalize foundational principles. At each stage, the LLM analyzes learner error patterns and biases to provide guidance for improvement, supporting the embodiment of practical judgment rather than mere knowledge transfer.

The first stage, \textit{Guided Recognition}, introduces learners to examples of high-quality and low-quality reviews, including those that violate Clarity through ambiguous language. Learners respond to questions such as “Does this review satisfy the criterion of Fairness?" to identify and internalize the characteristics of high-quality reviews. The LLM mentor provides explanations for each response to reinforce correct understanding of the criteria.

The second stage, \textit{Review Refinement Practice}, presents learners with review drafts that contain intentionally embedded flaws. Learners are tasked with revising and strengthening these drafts. Throughout this process, the LLM mentor offers targeted feedback such as “Your criticism remains ambiguous. Could you specify evidence indicating which experimental detail is problematic?" This guides learners to identify deficiencies and derive improved solutions on their own.

The final stage, \textit{Full Simulation}, requires learners to write a complete review of an entire paper. The LLM mentor analyzes the submitted draft according to the foundational principles and generates a comprehensive mentoring report. This report specifies concrete points of success and failure, for example, “Your review demonstrates strong Constructiveness, but it appears that you overlooked the content of Figure 5 in terms of Fidelity.” Such detailed feedback contributes to the embodiment of practical reviewing competence.

After completing all stages with LLM assistance, reviewers receive a \textit{Reviewer Certification}. This certification is not a mandatory qualification but rather serves as a positive signal indicating that the reviewer has voluntarily invested effort toward enhancing community quality and strengthening personal expertise. The LLM-assisted reviewer mentoring system provides a safe simulation environment that supports the systematic internalization of guidelines prior to actual evaluation, thereby establishing a sustainable and scalable foundation for enhancing reviewer capabilities.

\section{LLM-Assisted Reviewer Feedback System}

Even when reviewers have internalized core principles through prior training, the actual review process is far more complex and demanding than the training environment, making it difficult to guaranty that learned principles will be consistently applied. Junior reviewers, in particular, are prone to experiencing conceptual confusion or conflicts between principles when evaluating manuscripts. Senior reviewers may also commit errors due to time constraints or review fatigue. These practical limitations indicate that prior training alone cannot structurally mitigate the Reviewer Gap, suggesting the need for a mechanism that continuously supports principle-aligned decision-making during the review process itself. 

The LLM-assisted reviewer feedback system serves as a complementary mechanism designed to bridge precisely this training-execution gap. It automatically detects weaknesses in review drafts and supports optional refinement without imposing an additional burden on reviewers. By helping to ensure that principles internalized through prior training are maintained throughout the actual review process, this system functions as a connective mediator that secures continuity between education and execution, enabling learned competencies to be stably demonstrated in real evaluation contexts.

The system operates after a review draft is submitted and comprises structured stages powered by an LLM. In the first stage, \textit{Detection and Cross-Verification}, the LLM functions as a critic agent that meticulously scans the entire review text according to foundational principles and cross-verifies its alignment with the paper content. During this process, the LLM automatically identifies principle-violating elements such as factual errors, overlooked interpretations of the paper, and excessively definitive claims.

In the second stage, \textit{Evidence-based Feedback Generation}, the LLM focuses on specific review sentences requiring improvement, extracts relevant information from the paper, and provides concrete evidence and feedback to help reviewers justify their critiques more clearly. For example, if the system detects a Clarity-violating statement such as “The experiments on the dataset are insufficient,” it cross-references the paper to identify which datasets were used (e.g., ImageNet). It then generates feedback designed to strengthen the reviewer's reasoning and clarify the basis for their critique: “This comment requires improved Clarity. The paper uses ImageNet. If you believe additional experiments on a specific benchmark (e.g., Object365) are needed, please specify which additional benchmark you consider necessary and why.” This type of feedback encourages reviewers to articulate their critiques with greater precision and evidentiary support.

The generated feedback is not delivered immediately but first undergoes a final \textit{Reliability Testing} procedure~\cite{thakkar2025can, zhou2025benchmarking}. The LLM verifies that the feedback avoids issues such as inappropriate tones, reviewer-to-author phrasing, or uninformative praise. Only feedback that passes these checks is privately shared with reviewers and Area Chairs. Reviewers may then examine this feedback and revise their reviews accordingly. Revision is never mandatory; the system induces self-correction within a scope that does not increase reviewer burden. This optional-refinement structure respects the autonomy and expertise of reviewers while still offering a mechanism to address mistakes or principle violations.

The system offers tailored benefits for different reviewer groups. For junior reviewers, it facilitates an internalization process in which principles learned during prior training are reconfirmed and applied in actual review situations. For senior reviewers, it helps stabilize overall review quality by enabling pre-submission correction of overlooked points, misinterpretations, or ambiguous phrasing. Ultimately, this LLM-assisted approach is designed to systematically improve review quality while minimizing additional burden on reviewers~\cite{thakkar2025can}. By fostering clearer and more constructive exchanges between authors and reviewers, it is expected to contribute to a virtuous cycle that reinforces the robustness and overall health of the peer-review ecosystem.

\section{Discussion \& Alternative Views}
For the proposed LLM-assisted reviewer mentoring and feedback system to be continuously refined, it must not operate as a closed loop driven solely by automated analysis. A collaborative improvement framework involving human expert interaction is essential. Feedback generated by the system should be transparently provided not only to reviewers but also to expert groups such as Area Chairs, who can submit meta-feedback on validity, usefulness, and quality. Such expert verification addresses potential errors or biases in LLM-generated suggestions, and corrected cases can be accumulated as high-quality training data for continual refinement. This structure serves as a core mechanism enabling the system to become increasingly precise and to evolve in stable alignment with conference standards and scholarly values.

\begin{figure}[t]
    \centering
    \includegraphics[width=0.8\linewidth]{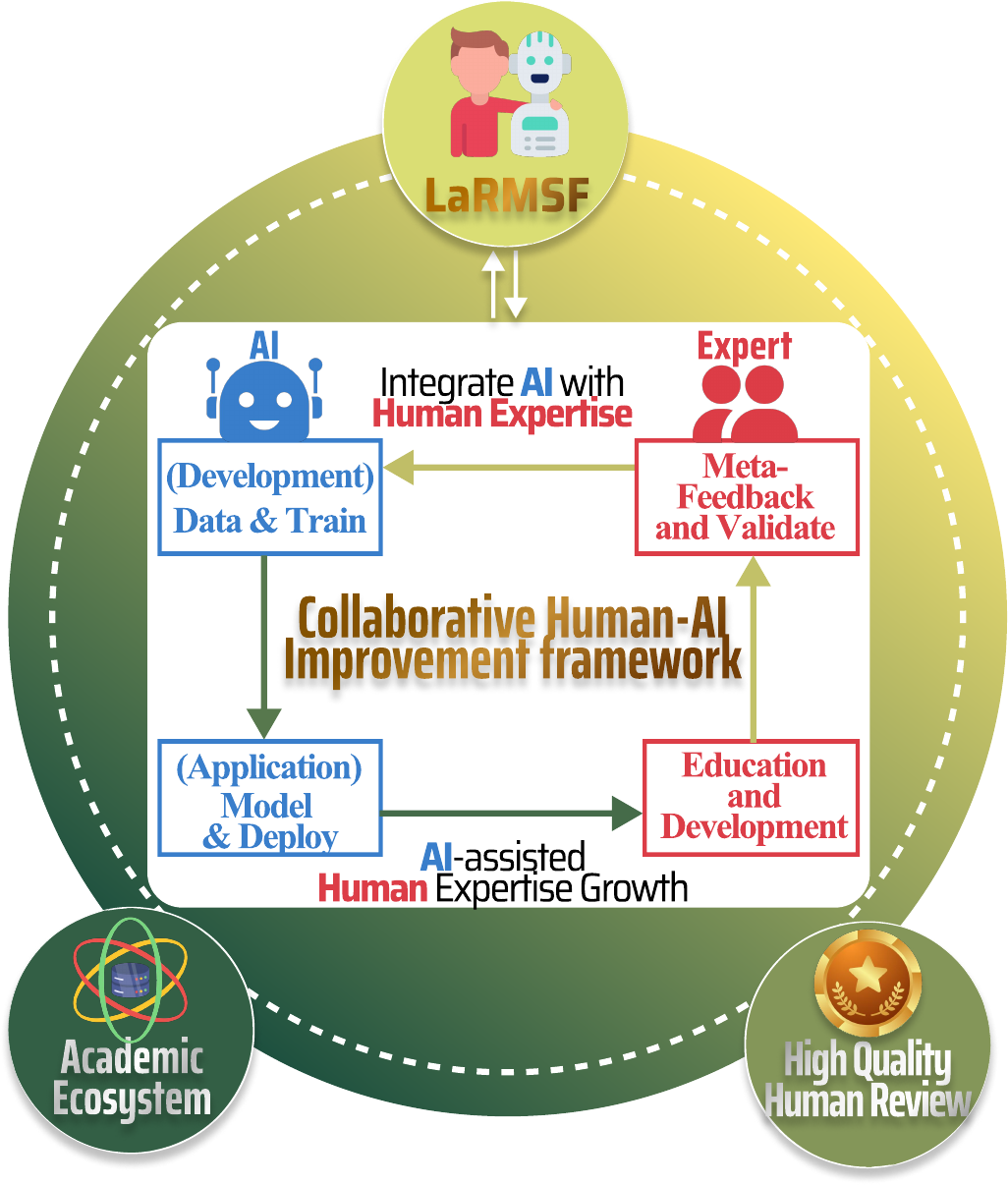}
    \caption{Macro-level virtuous cycle illustrating how improved reviews, refined research, and advanced AI technologies mutually reinforce the peer review ecosystem, supported by an internal collaborative human–AI improvement framework.}
    \label{fig:cycle}
\end{figure}

Beyond short-term efficiency gains, the LLM-assisted reviewer mentoring and feedback system has the potential to function as a structural mechanism that promotes the long-term advancement of scientific research. Reviews improved through system assistance provide authors with clearer and more constructive feedback, thereby substantively enhancing the rigor and completeness of individual research. The accumulation of research outputs that have undergone such rigorous peer review improves the overall reliability and reproducibility of scholarly literature, forming a robust knowledge foundation that drives scientific progress. Moreover, advanced research outcomes flow back as core resources for next-generation AI technology development, and these improved AI technologies can be reintegrated into peer review systems with enhanced analytical capabilities, such as detecting subtle logical leaps or more deeply understanding complex technical arguments. This macro-level virtuous cycle is expected to simultaneously drive the advancement of AI technology and the strengthening of scholarly community expertise, functioning as a foundational element of a sustainable academic ecosystem (Figure~\ref{fig:cycle}).

The proposed LLM-based system may face several criticisms and challenges. A primary concern is that excessive reliance on LLM feedback could undermine reviewers’ independent reasoning. If junior reviewers, in particular, accept LLM suggestions without critical examination, this could lead to long-term deskilling of reviewer capabilities. Critics opposing LLM utilization also point to the technical limitations of current models, including insufficient depth of technical insight and a tendency to  simply repeat or rephrase limitations already described by authors. Additionally, concerns exist that LLM mentors could provide biased feedback regarding specific research topics or methodologies, thereby amplifying academic biases and causing the homogenization of reviews.

These concerns are legitimate, yet the core philosophy of this paper lies in LLMs augmenting rather than automating human reviewers. We do not advocate for LLMs to completely replace human expert judgment. Rather, we argue that effectiveness is maximized when LLMs are utilized as educational and feedback systems that help human reviewers produce more consistent, high-quality reviews grounded in core principles. The risks of automation bias and deskilling are mitigated through the system's \textit{optional refinement} structure and explicit educational purpose. The system does not impose final judgments on reviewers but instead suggests potential errors and provides opportunities for self-correction, thereby aiming to train critical thinking abilities and strengthen competencies.

Concerns about bias amplification and homogenization can be addressed in the same context. The system does not homogenize reviewers' unique subjective opinions or critical perspectives. The LLM mentor does not evaluate the substantive validity of reviewers' judgments about specific research topics or methodologies; rather, it verifies whether the form in which those subjective opinions are expressed adheres to foundational principles. Consequently, the LLM does not amplify bias but instead assists reviewers in presenting their subjective critiques in clearer and fairer forms. Furthermore, the optional refinement structure ensures that reviewers retain full autonomy over their final assessments, preserving the diversity of perspectives that is essential to robust peer review.

We also acknowledge current limitations in the system's scope. The present framework primarily focuses on review text quality and does not yet incorporate an analysis of supplementary materials, such as code repositories or datasets. As AI-assisted review systems mature, extending coverage to the full submission package, including reproducibility verification, represents an important direction for future development. Such extensions would enable more comprehensive quality assurance while maintaining the human-centered augmentation philosophy proposed in this work.

\section{Conclusion}
The explosive growth of AI research has intensified the Reviewer Gap, threatening the sustainability of the peer review system. This position paper argues that the path forward lies not in automating reviewers but in augmenting their capabilities. We propose a dual framework combining LLM-assisted mentoring and feedback systems, grounded in foundational principles of high-quality reviews, to systematically strengthen reviewer expertise. This approach offers a practical pathway toward a more stable and trustworthy peer review process. Beyond immediate improvements, this work envisions a virtuous cycle:  high-quality reviews foster superior scientific advancements, and advances in AI further strengthen the review ecosystem. We hope this framework serves as a meaningful step toward building a sustainable and continuously improving scholarly community.

\section{Acknowledgments}
This work was supported by the Institute of Information \& Communications Technology Planning \& Evaluation (IITP) grant funded by the Korea government (MSIT) [RS-2021-II211341, Artificial Intelligence Graduate School Program (Chung-Ang University)] and by the National Research Foundation of Korea (NRF) grant funded by the Korea goverment (MSIT) (RS-2025-00556246).

\bibliography{aaai2026}

\end{document}